\pdfoutput=1

\documentclass[11pt]{article}

\usepackage[preprint]{acl}

\usepackage{times}
\usepackage{latexsym}
\usepackage{float}
\usepackage[T1]{fontenc}

\usepackage[utf8]{inputenc}

\usepackage{microtype}

\usepackage{inconsolata}

\usepackage{graphicx}
\usepackage{amsmath}
\usepackage{booktabs}
\usepackage{amsfonts}
\usepackage{cleveref}[capitalise]
\usepackage{tikz-cd,mathtools}
\usepackage{bm}
\renewcommand{\cref}{\Cref}

\title{SANDWiCH: Semantical Analysis of Neighbours for Disambiguating Words in Context ad Hoc}

\author{
Daniel Guzman-Olivares \\ Bulil Technologies \\ Autonomous University of Madrid\\\texttt{daniel.guzmano@estudiante.uam.es}
\And
Lara Quijano-Sanchez \\ Autonomous University of Madrid\\ \texttt{lara.quijano@uam.es}
\AND
Federico Liberatore \\ Cardiff University,\\ \texttt{liberatoref@cardiff.ac.uk}
}

\begin{document}
\maketitle
\begin{abstract}
The rise of generative chat-based Large Language Models (LLMs) over the past two years has spurred a race to develop systems that promise near-human conversational and reasoning experiences. However, recent studies indicate that the language understanding offered by these models remains limited and far from human-like performance, particularly in grasping the contextual meanings of words—an essential aspect of reasoning. In this paper, we present a simple yet computationally efficient framework for multilingual Word Sense Disambiguation (WSD). Our approach reframes the WSD task as a cluster discrimination analysis over a semantic network refined from BabelNet using group algebra. We validate our methodology across multiple WSD benchmarks, achieving a new state of the art for all languages and tasks, as well as in individual assessments by part of speech. Notably, our model significantly surpasses the performance of current alternatives, even in low-resource languages, while reducing the parameter count by 72\%.
\end{abstract}

\section{Introduction}
\label{sec:intro}
In 2022, OpenAI fine-tunned their previously released \textit{GPT-3} \cite{gpt3} model using Reinforcement Learning from Human Feedback (RLHF), resulting in the \textit{InstructGPT} model \cite{instructgpt}. Using this model and a massive dataset as a base, in November of the same year, OpenAI released a sibling model, the now famous \textit{ChatGPT}. The release served as the starting pistol for the ongoing race of chat-based Large Language Models. During the last two years, we have seen a consistent improvement in capabilities on different tasks \cite{minaee-sruvey, chatbot-arena} from bigger and newer models like \textit{LLama} \cite{llama}, PALM-2 \cite{anil2023palm2technicalreport} \textit{Falcon} \cite{falcon}, \textit{Mistral} \cite{mistral} or \textit{GPT-4} \cite{gpt4}. However, recent studies \cite{jack-of-all-trades, chatgpt-reason, leak, liu2023evaluating} suggest that these models struggle in logic reasoning tasks when the data is out of distribution from their train corpus and fail to match the performance of previously introduced specialized solutions. The difference in performance is particularly noticeable in the tasks requiring to assess the meaning in which words are used in a sentence, where we observe that recent chat-based models lag behind much smaller fine-tuned architectures \cite{eisenschlos-etal-2023-winodict, jack-of-all-trades, 10691283, qorib-etal-2024-decoder}.

The Word Sense Disambiguation (WSD) task consists in identifying the sense in which a word is used in some given context from a pool of possible senses \cite{wsd_survey}, e.g., in the sentence "\textit{The \underline{crane} was lifting a concrete block.}", a \textit{crane} refers to a lifting machine used in construction rather than a large, long-necked bird. Far from the massive chat-based language models that we can find today, the state-of-the-art models for this task are considerably smaller and were introduced during the last five years (e.g. \cite{consec, esc, bem, glossbert, kumar}) . In general, these models address the WSD problem by assessing the semantic similarity between the target word or its neighbouring context and the candidate definitions, using a fine-tuned encoder-based model \cite{consec, esc, bem, glossbert} or external resources \cite{bem, kumar}. Although these approaches generally provide remarkable results, they show a significant decrease in performance for verbs compared to other parts of speech, rare glosses, and underrepresented languages \cite{maru-etal-2022-nibbling, liu-liu-2023-ambiguity, consec}. These problematic cases suggest that current solutions struggle to accurately model word senses in low-resource settings, as verbs often have multiple possible senses that are unevenly distributed across different contexts, rare glosses are typically endemic to specialized domains, and training data for underrepresented languages is generally scarce. Addressing these deficiencies is crucial for improving generalization across out-of-distribution domains, where the traditional approach of training on large batches of annotated general-domain examples often fails to produce satisfying results \cite{maru-etal-2022-nibbling, bias} and bridging the WSD performance gap between English and low-resource languages. Ideally, a general solution to the WSD problem should reduce the dependency of performance on the frequency of senses and contexts present in the training data \cite{kilgarriff}, while addressing the issue at a structural level shared by all languages. Under this premise, we hypothesize that reframing the WSD task as a cluster discrimination task over a semantic network (e.g., BabelNet \cite{navigli-ponzetto-2010-babelnet}) could address the aforementioned challenges.

In this work, we introduce SANDWiCH\footnote{We release all the code for reproducing the paper results in \href{https://www.github.com/danielguzmanolivares/sandwich}{https://www.github.com/danielguzmanolivares/sandwich}}, a word disambiguation framework that leverages the close relationship between a candidate sense and its neighbors in a semantic network to shift the task from discriminating individual senses to discriminating semantically-close clusters. To this end, SANDWiCH incorporates additional elements into the two-level framework introduced in \citet{consec}, which consists of coarse sense retrieval followed by a fine-tuned encoder-based model. Specifically, we introduce the processing of the semantic network to ensure it is \textit{sense-separated} [\textbf{C1}], the inclusion of neighboring key concepts as part of the training data for the encoder-based models [\textbf{C2}], the separation of models by part of speech (POS) [\textbf{C3}], and the definition of a context-cluster score [\textbf{C4}].

Through extensive experimentation on the English all-words WSD task \cite{raganato-etal-2017-word}, we establish a new state of the art, achieving a 8\% improvement in F1 score across all datasets, consistently outperforming existing solutions in every subset, including those defined by individual datasets and parts of speech. We further evaluate our framework on the more challenging dataset introduced by \citet{maru-etal-2022-nibbling}, achieving an improvement over the previous state of the art ranging between 10-30\% depending on the dataset. Additionally, on the multilingual dataset \cite{Pasini_Raganato_Navigli_2021}, we improve state-of-the-art results for all languages, with particularly notable gains in underrepresented ones.

Therefore, the key contributions of this work are as follows:
\begin{itemize}
    \item \textbf{SANDWiCH framework}: We introduce a novel word sense disambiguation framework that shifts the focus from individual sense discrimination to cluster-based sense discrimination, utilizing sense-separated semantic networks and neighboring key concepts to improve performance and robustness.

    \item \textbf{State-of-the-art results on English datasets}: Our system achieves a 8\% improvement in F1 score on the English all-words WSD task, consistently surpassing the state of the art across all datasets and parts of speech, including the challenging dataset introduced by \citet{maru-etal-2022-nibbling}.

    \item \textbf{Multilingual generalization}: The proposed framework generalizes effectively to multilingual settings, achieving state-of-the-art results across all languages in the multilingual WSD dataset \cite{Pasini_Raganato_Navigli_2021}, with significant improvements in underrepresented languages.
    
\end{itemize}

\section{Related Work}
Historically, the WSD problem was introduced in the second half of the twentieth century as part of machine translation efforts \cite{weaver, BarHillel1960ThePS}, later evolving into a standalone problem. Early successful approaches primarily relied on rule-based algorithms, statistical methods, and unsupervised techniques \cite{Gale1992AMF, Yarowsky1992WordSenseDU, lesk, cowie-etal-1992-lexical-disambiguation, Yarowsky1995UnsupervisedWS}. The development of large-scale structured language resources like Wikipedia and BabelNet \cite{navigli-ponzetto-2010-babelnet} enabled models to use gloss similarity heuristics and graph proximity metrics to address the WSD problem \cite{moro-etal-2014-entity, wang-etal-2015-sense, mccarthy-etal-2016-word, fuzzy_hindi}.

The introduction of the first word embedding algorithms (e.g., Word2Vec \cite{word2vec}, fastText \cite{fasttext}, or GloVe \cite{glove}) significantly advanced WSD performance by leveraging seq-to-seq supervised approaches \cite{Kgebck2016WordSD, taghipour-ng-2015-semi, yuan-etal-2016-semi, luo-etal-2018-incorporating}. The use of word embeddings as a foundation for neural approaches led to substantial performance gains, which became even more pronounced with the adoption of dynamic embeddings from encoder models (e.g., BERT \cite{bert}, RoBERTa \cite{liu2019robertarobustlyoptimizedbert}, or DeBERTa \cite{deberta}), derived from the transformer architecture \cite{attention}. Fine-tuning an encoder model has since then become the cornerstone of top-performing systems, which can be broadly categorized into two variants based on their approach to the WSD problem.

The first variant comprises purely transformer-based architectures that leverage the representational power of large encoder models. These models often frame the problem by jointly encoding all candidate definitions alongside the given sentence to extract the correct sense of a target word \cite{scarlini-etal-2020-contexts, glossbert, hadiwinoto-etal-2019-improved}. Notable examples include ConSec \cite{consec}, the previous state-of-the-art, which encodes not only the candidate senses of the target word but also non-ambiguous or already disambiguated words in the sentence; BEM \cite{bem}, which separates the encoding of glosses and context to compute similarity through a dot product; and ESC \cite{esc}, which redefines the WSD problem as a span extraction task, analyzing the concatenation of all possible senses to determine the start and end indices of the correct sense.

The second variant includes transformer models that integrate external information, usually from a lexical knowledge base or semantic network \cite{loureiro-jorge-2019-language, conia-navigli-2021-framing, song-etal-2021-improved-word}. Notable examples include EWISE \cite{kumar} and its improved version EWISER \cite{ewiser}, which incorporate WordNet information into the neural model; DHFM \cite{dhfm}, which enriches pretrained embeddings with graph encodings of senses; and \citet{mizuki-okazaki-2023-semantic}, which use synonyms and hypernyms from WordNet to train an encoder via a triplet loss over semantically related glosses. Recent approaches have started exploring parallel alternatives, such as \citet{dong-sifa-2024-word}, which propose using neurosymbolic embeddings that reach 90\% F1 for target senses with an explicit class structure (about 70\% of the \citet{raganato-etal-2017-word} dataset), and \citet{zhang}, which use a representation based on superposition states to eliminate dependency on training set size and improve accuracy for rare senses.

Although both variants produce competitive results, to the best of our knowledge, no system has surpassed the 82\% F1 score on the unified benchmark \cite{raganato-etal-2017-word}. Additionally, \citet{maru-etal-2022-nibbling} highlighted that generalization to out-of-domain tasks remains a challenge, complicating the ability of current solutions to scale to specialized domains. Moreover, most modern systems rely heavily on encoder models predominantly trained in English, limiting their applicability to underrepresented languages \cite{consec}. To address these challenges, we propose reframing the WSD problem as a semantic cluster discrimination task within a semantic network (BabelNet) and, in the next section, introduce the SANDWiCH framework as a comprehensive solution to the multilingual word sense disambiguation problem.

\section{The SANDWiCH framework}
\subsection{Theoretical motivation}
\label{sec:theoretical}
\begin{figure*}[ht]
    \centering
    \includegraphics[scale=.25]{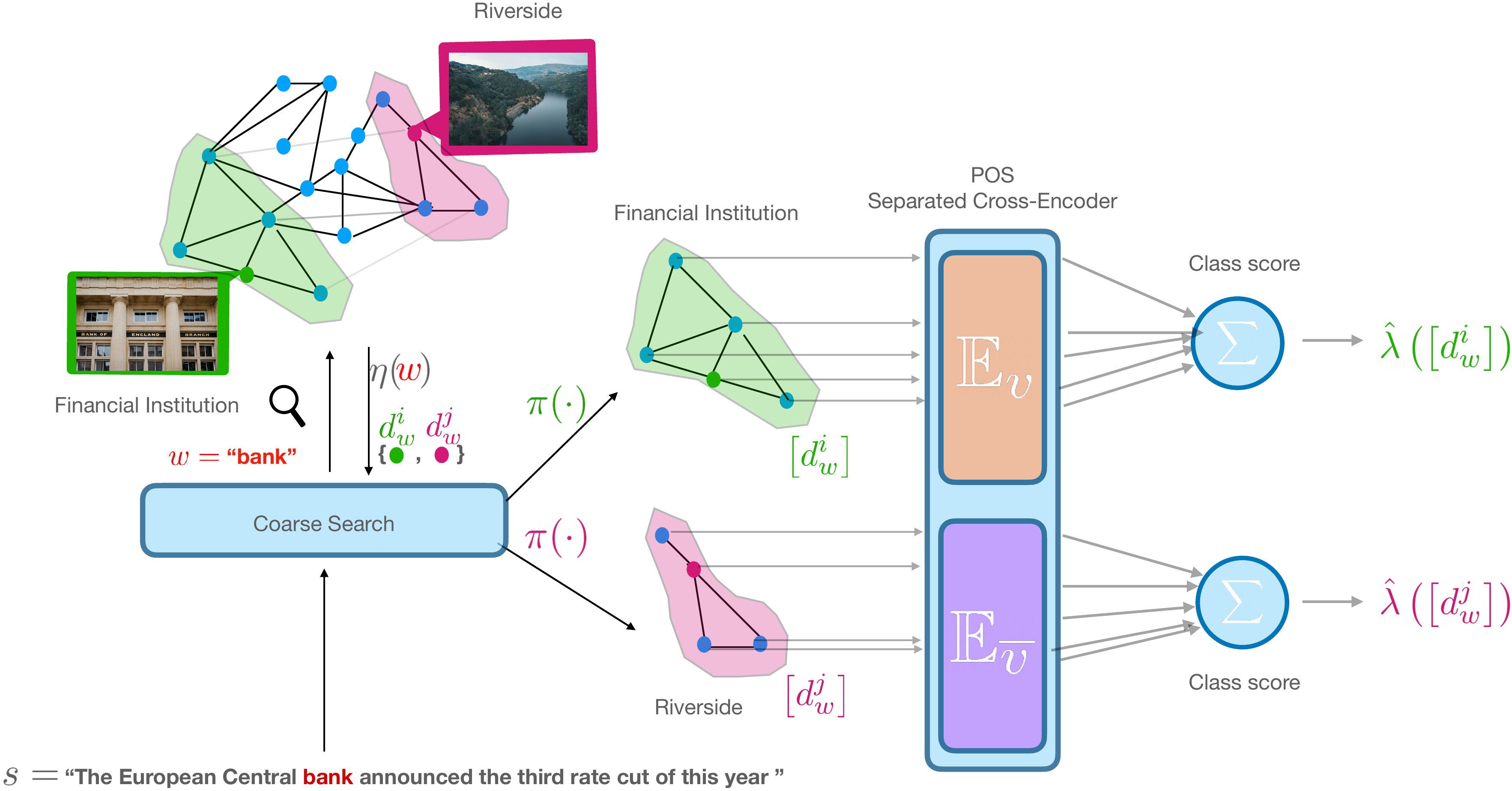}
    \caption{Illustration of the SANDWiCH architecture in the processing of the word \textit{bank} in context.}
    \label{fig:architecture}
\end{figure*}
Formally, a written language $\mathcal{L}$, can be defined by the generator $\mathcal{L} := <\mathcal{V}, \bigoplus>$, where $\mathcal{V}$ is a vocabulary and $\bigoplus$ is the word concatenation operation. Using this notation we denote the dictionary space, that contains the definitions of every word sense as $\mathcal{D} \subset \mathcal{L}$. Naturally, we can define a function $\eta: \mathcal{V} \rightarrow \mathcal{P}(\mathcal{D})$, mapping each word to a set of possible definitions. Additionally, given a sentence $s \in \mathcal{L}$, and a target word $w \in s$, we can define a function $\varphi$ that selects the correct definition from $\eta(w)$. The disambiguation process can be then formalized as:
\[
\begin{aligned}
\varphi \circ \eta: \mathcal{V} \times \mathcal{L} &\xrightarrow{\eta} \mathcal{P}(D) \times \mathcal{V} \times \mathcal{L}  \xrightarrow{\varphi} \mathcal{D} \\
(w, s) &\mapsto \left( \{d_w^i\}_{i=1}^n, w, s\right) \mapsto d_w^k
\end{aligned}
\]
Where $d_w^i$ is a definition associated with the word $w$, $n$ is the total possible definitions associated with $w$, and $d_w^k$ is the correct definition for $w$ in $s$. Usually $\eta$ is provided and the WSD task consists in approximating $\varphi$.

The SANDWiCH framework assumes that we are additionally given a graph structure $G := (\mathcal{D}, \mathcal{E})$ over the definitions space $\mathcal{D}$, in which an edge $(d_i, d_j) \in \mathcal{E} \subset \mathcal{D} \times \mathcal{D}$ connecting the definitions $d_i, d_j \in \mathcal{D}$ exists if $d_i$ is semantically related with $d_j$ (e.g. \textit{apple (Fruit.)} $\sim$ \textit{fruit (The ripened reproductive body of a seed plant.))}. Using this notation, we can define the sense neighbourhood of $d_i \in \mathcal{D}$ as $\mathcal{N}(d_i) := \{d_j : (d_i, d_j) \in \mathcal{E}\}$, and the union set of all neighbourhoods of candidate definitions given a word $w$ as $\mathcal{D}_{\eta(w)} := \bigcup_{d_w^i \in \eta(w)}\{\mathcal{N}(d_w^i)\}$.

The graph that SANDWiCH uses is a modified version of the semantic graph to ensure that the graph is \textit{sense-separable}:

\textbf{Sense-Separability Condition:} \textit{Given the semantic graph} $G := (\mathcal{D}, \mathcal{E})$, \textit{defined as above, we say that the graph is sense-separable if and only if for any given word} $w$ \textit{we have that}:
\begin{equation*}
    \forall d_w^i, d_w^j, \in \eta(w):  \mathcal{N}(d_w^i) \cap \mathcal{N}(d_w^j) = \emptyset
\end{equation*}
In practice, its enough with eliminating the edges connecting neighborhoods in the graph $G$. Given a word $w$ if the sense-separability condition holds, and we consider the subgraph defined by $\mathcal{D}_{\eta(w)}$ and its connecting edges $\mathcal{E}_{\eta(w)}$, the relation $\sim_w : d_i \sim_w d_j \iff \exists k: d_i,d_j \in \mathcal{N}(d_w^k \subset \mathcal{D}_{\eta(w)})$ defines an equivalence relation over $\mathcal{D}_{\eta(w)}$ and therefore we have that:
\\\\
\begin{tikzcd}[column sep=small, row sep=large]
    \mathcal{P}(\mathcal{D}) \times \mathcal{V} \times \mathcal{L} \arrow[r, "\varphi"] \arrow[d, "\bigcup"] & \eta(w) \\
    \mathcal{D}_{\eta(w)} \times \mathcal{V} \times \mathcal{L} \arrow[r, "\pi"] \arrow[ru, "\varphi^{*}"] & \left( \mathcal{D}_{\eta(w)} / \sim_w \times \mathcal{V} \times \mathcal{L} \right) \arrow[u, "\lambda"]
\end{tikzcd}
\\\\
Where $\varphi^*$ is the natural extension of $\varphi$ to the $\mathcal{D}_{\eta(w)}$ domain (i.e. $\varphi^*(d_w^i) = d_w^i$ if $\varphi\left( (d_w^{(1)}, \dots, d_w^{(n)}), w, s \right) = d_w^i$, otherwise $\varphi^*(d_w^i) = \emptyset$); $\pi$ is the canonical projection of the equivalence relation and $\lambda$ maps an equivalence class $\left[d_w^i\right]$ to the correct definition. Since $\varphi$ is class invariant under $\sim_w$, then $\varphi \cong \lambda \circ \pi$, and there exists a unique $\lambda$ satisfying this relation \cite{key0214415m}. This means that we can approximate the disambiguation process given by $\varphi$ at the definition level by the equivalence class of all semantically related definitions in the graph $G_{\eta(w)} = (\mathcal{D}_{\eta(w)}, \mathcal{E}_{\eta(w)})$. To this end, we use the following approximation:

\begin{equation*}
    \hat{\lambda} = \arg\max_{[d_w^i] \in \frac{\mathcal{D}_{\eta(w)}}{\sim_w}} \sum_{d_w^j \in [d_w^i]} \left( \mathbb{E}_{v}(d_w^j) + \mathbb{E}_{\overline{v}}(d_w^j) \right) \delta_j
\end{equation*}

Where $\mathbb{E}_{\overline{v}}\left(\cdot\right)$ is an encoder-based model finetuned using data including every POS except verbs to predict the probability of a definition $d_w^j$ being semantically relevant given the word $w$ in a given context. Analogously, we can define  $\mathbb{E}_v(\cdot)$ for an encoder-based model using data including nouns and verbs only. Finally, the weight scores $\delta_j$ are defined as
\begin{equation*}
    \delta_{j} = \frac{e^{2|\mathbb{E}_{v}(d_w^j) + \mathbb{E}_{\overline{v}}(d_w^j) - 1|}}{\sum_{d_w^j \in [d_w^i]} e^{2|\mathbb{E}_{v}(d_w^j) + \mathbb{E}_{\overline{v}}(d_w^j) - 1|}}
\end{equation*}

\subsection{Implementation Details}
\label{sec:implementation}
From a theoretical perspective, the Word Sense Disambiguation (WSD) task consists of two main components. The first is an information retrieval step, where the goal is to estimate $\eta$, the top-K sense candidates associated with a given word. For certain parts of speech, such as adjectives and adverbs, this retrieval may not be necessary due to their limited number of possible senses. However, for nouns and especially verbs, which often have a wide range of senses, this step is critical to narrowing down the candidate definitions and ensuring a manageable input size for the disambiguation process.

In the SANDWiCH framework, this retrieval is managed by a coarse sense retrieval module (see \cref{fig:architecture}), which fine-tunes a DeBERTa-v3-xsmall model as a cross-encoder \cite{reimers-gurevych-2019-sentence} to estimate the relevance of a candidate definition given the sentence and target word. We train this model using the SemCor corpus \cite{miller} combined with datasets from \citet{raganato-etal-2017-word}, following established methods to classify candidate definitions as relevant or not. The top-K candidate senses are ranked by probability, and we set K=30 as per \citet{consec}, achieving a recall of 98-99\% across all datasets.

The second step selects the most appropriate definition from the retrieved candidates. Instead of directly estimating $\varphi$, we focus on composing equivalence classes through $\lambda \circ \pi$, reducing reliance on specific word-level data by estimating at the equivalence class level. For this approach to work, the sense-separability condition must hold, meaning the semantic clusters in the graph must be disjoint. We extract the sense graph from BabelNet \cite{navigli-ponzetto-2010-babelnet} and remove edges connecting senses of the same word to ensure clean separability. The equivalence classes are defined as the immediate neighborhoods of the target word's senses.

Initial experiments revealed a significant performance boost by partitioning the training data into two groups: one for nouns and verbs, and another for nouns, adjectives, and adverbs. Training separate cross-encoders for these groups further enhanced performance, even beyond a standard ensemble of models, as discussed in \cref{sec:ablation}.

For training, we generate positive and negative examples by sampling from the neighborhoods of correct and incorrect senses. Unlike the coarse retrieval step, all elements within a neighborhood share the same label. The input consists of a concatenated sentence-definition pair $(s, d_w^i)$, where the word $w$ in sentence $s$ is marked with special tokens [d]. We use DeBERTa-v3-small as the backbone model for the cross-encoders, training with a batch size of 64, 10 epochs, a learning rate of $2e^{-5}$, and gradient clipping at 1. A cosine annealing scheduler \cite{loshchilov2017sgdrstochasticgradientdescent} and binary cross-entropy with logits are used as optimization methods.

After training, the class score is computed using the formula outlined in \cref{sec:theoretical}. The $\delta_{ij}$ weights are derived from the softmax of the absolute difference between the model’s predictions for relevance and non-relevance, which represents its confidence in assigning the correct sense cluster. Additional training details can be found in \cref{apdx:training}

\section{Experimentation}
\label{sec:experiments}
In this section, we present and discuss the results of our experiments to evaluate the SANDWiCH framework against existing alternatives. In \cref{sec:all-wsd}, we first assess our model's performance on the English all-words benchmark \cite{raganato-etal-2017-word}, breaking down results by individual datasets and parts of speech. Following this, in \cref{sec:more_difficult}, we examine how well SANDWiCH generalizes to previously unseen domains and rare senses, using the more challenging dataset from \citet{maru-etal-2022-nibbling}, and compare it to the current state of the art. We then perform an ablation study in \cref{sec:ablation} to evaluate the individual contribution of each system component. In \cref{sec:multilingual}, we investigate the framework’s adaptability to other languages. Finally, in \cref{sec:efficient}, we explore alternative backbone models for the cross-encoders and analyze the trade-off between model size and performance.

\subsection{All-words English WSD}
Introduced in 2017, the all-words English WSD benchmark is the most widely used standard for evaluating WSD systems. It comprises five datasets: Senseval-2002 (\textbf{SE2}) \cite{edmonds-cotton-2001-senseval}, Senseval-2003 (\textbf{SE3}) \cite{snyder-palmer-2004-english}, Semeval-2007 (\textbf{SE7}) \cite{pradhan-etal-2007-semeval}, Semeval-2013 (\textbf{SE13}) \cite{navigli-etal-2013-semeval}, and Semeval-2015 (\textbf{SE15}) \cite{moro-navigli-2015-semeval}. Following prior work \cite{raganato-etal-2017-neural, glossbert, bem, consec}, we use Semeval-2007 as the development set and train on the SemCor corpus. Our results, reported by individual dataset and POS, are summarized in \cref{table:all-wsd}.

The SANDWiCH framework significantly outperforms previous state-of-the-art methods across all datasets and parts of speech, improving the overall F1 score by seven points. Notably, noun disambiguation sees an eight-point increase, highlighting the effectiveness of the equivalence class approximation for the WSD task.

\begin{table*}[ht]
\centering
\begin{tabular}{lcccccccccc}
\toprule
\textbf{Model} & \textbf{SE07} & \textbf{SE2} & \textbf{SE3} & \textbf{SE13} & \textbf{SE15} & \textbf{Nouns} & \textbf{Verbs} & \textbf{Adj.} & \textbf{Adv.} & \textbf{ALL} \\ 
\midrule
MFS - SemCor & 54.5 & 65.6 & 66.0 & 63.8 & 67.1 & 67.7 & 49.8 & 73.1 & 80.5 & 65.5 \\ 
BERT(base) & 68.6 & 75.9 & 74.4 & 70.6 & 75.2 & 75.7 & 63.7 & 78.0 & 85.8 & 73.7 \\ 
SVC - Ensemble & 69.5 & 77.5 & 77.4 & 76.0 & 78.3 & 79.6 & 65.9 & 79.5 & 85.5 & 76.7 \\ 
GlossBERT & 72.5 & 77.7 & 75.2 & 76.1 & 80.4 & 79.8 & 67.1 & 79.6 & 87.4 & 77.0 \\ 
ARES & 71.0 & 78.0 & 77.1 & 78.7 & 75.0 & 80.6 & 68.3 & 80.5 & 83.5 & 77.9 \\ 
EWISER & 71.0 & 78.9 & 78.4 & 78.9 & 79.3 & 81.7 & 66.3 & 81.2 & 85.8 & 78.3 \\ 
WMLC & 72.2 & 78.4 & 77.8 & 76.7 & 78.2 & 80.1 & 67.0 & 80.5 & 86.2 & 77.6 \\ 
BEM & 74.5 & 79.4 & 77.4 & 79.7 & 81.7 & 81.4 & 68.5 & 83.0 & 87.9 & 79.0 \\ 
ESCHER & 76.3 & 81.7 & 77.8 & 82.2 & 83.2 & 83.9 & 69.3 & 83.8 & 86.7 & 80.7 \\ 
CoNSeC & 77.4 & 82.3 & 79.9 & 83.2 & 85.2 & 85.4 & 70.8 & 84.0 & 87.3 & 82.0 \\
QR-WSD & 74.5 & 80.6 & 79.1 & 80.0 & 84.7 & 83.7 & 71.4 & 82.8 & 86.7 & 80.5 \\
GPT4o & - & 76.3 & 73.2 & 79.7 & 83.7 & 81.2 & 66.3 & 79.0 & 71.3 & 77.4 \\
GPT4 & - & 74.3 & 70.0 & 77.4 & 79.5 & 78.6 & 59.7 & 79.5 & 74.0 & 74.6 \\
GPT-3.5 & - & 63.1 & 59.2 & 63.8 & 70.5 & 68.1 & 46.7 & 66.6 & 64.8 & 63.3 \\
SANDWiCH & \textbf{81.2} & \textbf{88.5} & \textbf{84.9} & \textbf{92.5} & \textbf{91.7} & \textbf{94.0} & \textbf{74.6} & \textbf{86.8} & \textbf{91.6} & \textbf{89.0} \\ 
\bottomrule
\end{tabular}
\caption{Performance (F1 score) of various models, broken down by task and POS in the all-words English WSD benchmark. The best results are highlighted in bold. The compared systems include MFS, which selects the most common sense from SemCor, BERT base \cite{bert}, SVC \cite{vial-etal-2019-sense}, GlossBERT \cite{glossbert}, ARES \cite{scarlini-etal-2020-contexts}, EWISER \cite{ewiser}, WMLC \cite{conia-navigli-2021-framing}, BEM \cite{bem}, ESCHER \cite{esc}, ConSeC \cite{consec}, QR-WSD \cite{zhang}, GPT4o, GPT4, and GPT-3.5. are the most recent versions of the \href{https://chatgpt.com/}{ChatGPT} model available at the time of writing.}
\label{table:all-wsd}
\end{table*}

\label{sec:all-wsd}
\subsection{A More Challenging Dataset}
\label{sec:more_difficult}
In this experiment, we reproduce and evaluate the performance of the SANDWiCH framework on the rare senses benchmark introduced by \citet{maru-etal-2022-nibbling}. This benchmark consists of four parts: a dataset designed to test WSD systems on rare and out-of-domain senses (\textbf{42D}); a collection of the most frequent errors made by state-of-the-art models on the all-words English WSD benchmark (\textbf{hardEN}); a WSD task similar in nature to those found in the all-words English WSD benchmark (\textbf{S10}); and (\textbf{softEN}), which is the opposite of the hardEN dataset.

We present the results of previous reported systems alongside the ConSeC model, which represented the state of the art in the all-words English WSD benchmark, in \cref{table:difficult}. Notably, SANDWiCH significantly outperforms all models, achieving improvements of over 10 F1 points in S10, 22 F1 points in 42D, 45 F1 points in hardEN, and two F1 points in softEN.

\begin{table*}[ht]
    \centering
    \begin{tabular}{lcccccccccc}
        \hline
        \textbf{\#Dataset} & \textbf{ARES} & \textbf{BEM} & \textbf{ESC} & \textbf{EWS} & \textbf{GEN} & \textbf{GBT} & \textbf{SYN} & \textbf{CSC} & \textbf{SandWiCH}\\
        \hline
        S10     & 77.9 & 77.1 & 78.0 & 76.1 & 72.3 & 75.8 & 64.0 & 77.5 & \textbf{87.5}\\
        42D     & 41.8 & 53.2 & 58.9 & 43.9 & 50.2 & 45.7 & 32.8 & 56.6 & \textbf{77.1}\\
        softEN  & 78.7 & 80.3 & 83.7 & 79.2 & 76.4 & 77.1 & 63.4 & 87.7 & \textbf{89.4}\\
        hardEN  & 0.0  & 0.0  & 0.0  & 0.0  & 0.0  & 0.0  & 0.0  & 7.35 & \textbf{53.4}\\
        \hline
    \end{tabular}
    \caption{F1 performance metrics on the \citet{maru-etal-2022-nibbling} benchmark. The compared models are ARES \cite{scarlini-etal-2020-contexts}, BEM \cite{bem}, ESC \cite{esc}, EWS \cite{ewiser}, GEN \cite{bevilacqua-etal-2020-generationary}, GBT \cite{glossbert}, SYN \cite{scozzafava-etal-2020-personalized}, and CSC\cite{consec}. Best scores are highlighted in bold.}
    \label{table:difficult}
\end{table*}

\subsection{Ablation Study}
\label{sec:ablation}
In this section, we assess the individual contributions of each component within the SANDWiCH framework to better understand their interrelations. To do this, we first explore the practical and theoretical contributions that enable the architecture to function effectively. As introduced in \cref{sec:intro}, the three main pillars supporting the SANDWiCH framework are: the use of equivalence classes instead of single senses, the sense-separability condition in the semantic graph, and the part-of-speech (POS) separation of the cross-encoders for computing class scores.

In this ablation study, we first analyze the effect of using equivalence classes instead of senses directly, observing the expected decrease in performance. This decline occurs because the system loses robustness against the frequency bias in the training data \cite{maru-etal-2022-nibbling, bias}, making it overly dependent on the training distribution and limiting its ability to generalize beyond the training domain.

If we maintain the use of classes but cannot ensure the semantic graph is sense-separable, we introduce noise into the training set, particularly with word-definition pairs labeled both positively and negatively for the same sentence.

Finally, we find that employing POS separation in the cross-encoders leads to a considerable performance increase compared to using a standard ensemble of two cross-encoders trained on the entire dataset. This gain may stem from the differing disambiguation strategies for each POS: verbs typically rely on objects, subjects, actions, and tense information \cite{hashimoto-tsuruoka-2015-learning, Wagner2009VerbSD}, while nouns, adjectives, and adverbs focus more on their interrelations \cite{nouns}. All results are reported in \cref{table:ablation} in which we evaluate the performance in the all-words English dataset in the aforementioned cases.

\begin{table}[]
\begin{tabular}{lc}
\hline
\multicolumn{1}{c}{{\textbf{Active Components}}} & {\textbf{Score ALL}} \\ \hline
Classess                                       & 53.5                     \\
Classes + Encoders                    & 66.6                     \\
Classes + Separability                     & 79.5                     \\
Encoders + Separability               & 57.5                     \\
Classes + Encoders (No sep.)     & 55.1    \\
Whole Pipeline                                       & 89.0                     \\ \hline
\end{tabular}
\caption{Ablation study on different components of the SANDWiCH pipeline, \textit{Classes} denotes using the equivalence class structure instead of senses directly, \textit{Encoders} refers to the splitting of the cross-encoders by POS as described in \cref{sec:theoretical}, \textit{Separability} is whether the separability condition holds or not, and \textit{Encoders (No sep.)} is an ensemble of cross-encoder non-separted by POS. The F1 ALL score refers to the score in the all-words English dataset.}
\label{table:ablation}
\end{table}

\subsection{Multilingual WSD}
\label{sec:multilingual}
In this setting, we explore the adaptability of the SANDWiCH pipeline to other languages. Specifically, we assess the performance of our solution across nine languages in the XL-WSD dataset \cite{Pasini_Raganato_Navigli_2021}. Since the DeBERTa-v3 model is trained exclusively in English, we use mBART-50 \cite{mbart} as the backbone model. For each language, we also adapt the BabelNet semantic network to ensure it meets the sense-separability assumption.

In this context, SANDWiCH outperforms the current state of the art in every tested language. The improvements are consistent across all language groups, with gains exceeding nine F1 points in Germanic languages (English, German, and Dutch), eight F1 points in Romance languages (Spanish, Italian, and French), 18 F1 points in Finno-Ugric languages (Estonian), and 25 F1 points in Japonic languages (Japanese).

Additionally, we evaluated our system in Croatian, representing a low-resource language in the Slavic language group, achieving a competitive performance of 84.1 F1 points. For the other low-resource languages (Estonian, Dutch, and Japanese) in our tests, SANDWiCH's reduced dependency on individual senses resulted in the most significant improvements, surpassing the next best approach by over 20 F1 points.

\begin{table*}[ht]
    \centering
    \begin{tabular}{lccccc}
        \hline
        \textbf{Language} & \textbf{SyntagRank} & \textbf{EWISER} & \textbf{XLMR} & \textbf{ConSeC} & \textbf{SANDWiCH} \\
        \hline
        English    & 70.0 & 73.3 & 76.3 & 79.0 &  \textbf{88.9}\\
        Dutch      & 56.0 & 57.5 & 59.2 & 63.3 &  \textbf{83.7}\\
        Estonian   & 56.3 & 66.0 & 66.1 & 69.8 &  \textbf{89.5}\\
        French     & 70.0 & 80.9 & 83.9 & 84.4 &  \textbf{92.8}\\
        German     & 76.0 & 80.9 & 83.1 & 84.2 &  \textbf{93.2}\\
        Italian    & 69.6 & 74.6 & 77.6 & 79.3 &  \textbf{86.6}\\
        Japanese   & 57.5 & 55.8 & 61.9 & 63.0 &  \textbf{85.7}\\
        Spanish    & 68.6 & 71.9 & 75.9 & 77.4 &  \textbf{84.0}\\
        Croatian   & -    & -    & -    & -    &  \textbf{84.1}\\
        \hline
    \end{tabular}
    \caption{Comparison of F1 scores across different languages in the XL-WSD \cite{Pasini_Raganato_Navigli_2021} for SyntagRank \cite{scozzafava-etal-2020-personalized}, EWISER \cite{ewiser}, XLMR \cite{Pasini_Raganato_Navigli_2021}, ConSeC \cite{consec}, and SANDWiCH.}
    \label{table:multilingual}
\end{table*}
\subsection{Backbone model efficiency analysis}
\label{sec:efficient}
As mentioned in \cref{sec:implementation}, we use DeBERTa-v3-small as the backbone system in the SANDWiCH architecture. To evaluate the trade-off between parameter count and performance, we modify the backbone model and analyze its impact on the all-words English WSD dataset. In \cref{table:efficiency}, we compare several models: BERT-base, BERT-large, BART-large \cite{lewis-etal-2020-bart}, RoBERTa-base, RoBERTa-large, DeBERTa-xsmall, DeBERTa-small, DeBERTa-base, and DeBERTa-large. Our results indicate that the DeBERTa family offers the highest overall performance, with performance gains diminishing as model size increases. For instance, the leap from DeBERTa-xsmall to DeBERTa-large (with a 1300\% increase in parameter count) yields substantial improvements (4.5 F1 points), but moving from DeBERTa-small to DeBERTa-large results in only a 0.1 F1 point gain. This suggests that DeBERTa-small provides the optimal balance between parameter count and performance, outperforming the ConSec model by six F1 points while using just 28\% of its parameters.

\begin{table}[]
\begin{tabular}{ccc}
\hline
\textbf{Model}    & \textbf{N° Params.} & \textbf{F1 Score} \\ \hline
DeBERTa v3 xsmall & 22M                    & 84.6                    \\
DeBERTa v3 small  & 44M                    & 89.0                    \\
DeBERTa v3 base   & 86M                    & 89.0                    \\
DeBERTa v3 large  & 304M                   & 89.1                    \\
BERT base         & 110M                   & 78.7                    \\
BERT large        & 340M                   & 80.2                    \\
BART large        & 406M                   & 83.3                    \\
RoBERTa base      & 125M                   & 80.5                    \\
RoBERTa large     & 355M                   & 81.9\\                   
\hline
\end{tabular}
\caption{Performance of the SANDWiCH framework in the all-words English dataset \cite{raganato-etal-2017-word} changing the backbone model.}
\label{table:efficiency}
\end{table}
\section{Results Analysis}
The performance of the proposed SANDWiCH framework across existing datasets demonstrates that reframing the WSD problem as a discrimination task over semantically related clusters effectively addresses the limitations of current solutions, confirming our initial hypothesis. Specifically, in the all-words English benchmark, we surpass the previous state of the art across each dataset and in the combined total (ALL). This improvement extends to rare senses and out-of-domain data, as shown by results on the \citet{maru-etal-2022-nibbling} dataset, where SANDWiCH significantly outperforms prior solutions. This success includes cases where words have a large number of possible senses (see \cref{apdx:polysemic}), indicating that our approach mitigates challenges in such scenarios.

Additionally, we analyze the individual contributions of each component in the architecture, concluding that the key to SANDWiCH's success is the creation of separable clusters over the semantic network. Furthermore, separating cross-encoders by POS leads to considerable performance gains. We also extend the framework to multiple languages, outperforming all existing alternatives and making significant strides in low-resource language disambiguation. Notably, we demonstrate that SANDWiCH achieves these results with only 28\% of the parameters used by the previous state-of-the-art, proving the robustness of the sense-cluster approach.

\section{Conclusion}
In this paper, we introduced the SANDWiCH framework, a novel approach to the WSD problem, which arose from the hypothesis that reframing the disambiguation task as sense cluster discrimination over a semantic network could address the challenges faced by previous state-of-the-art solutions when generalizing to low-resource languages and domains. Through extensive experimentation, we confirmed our hypothesis, surpassing the state of the art across all benchmarks, including rare senses and multiple languages. Furthermore, we evaluated various alternatives for the backbone model and demonstrated the efficiency of our architecture, achieving a 72\% reduction in model size while still surpassing the state of the art.

In future work, it would be valuable to explore alternative methods for creating sense clusters, extend our approach to additional languages, and investigate whether SANDWiCH’s disambiguation capabilities can serve as a baseline or be combined with existing solutions for translation into low-resource languages and specialized text analysis.

\section*{Limitations}
As detailed in \cref{sec:implementation}, the implementation of the SANDWiCH framework requires of a previously given semantic network. However depending on the language, this might be a complicated resource to get or not as complete as needed for ensuring a reasonable accuracy (e.g. for low-resource languages). Our architecture also depends on the performance of the cross-encoders used to calculate the score of the equivalence classes, even if we manage to greatly improve the performance for some underrepresented languages, the backbone models used are not available for every language and that can limit the usability of our proposed solution.

\section*{Acknowledgments}
We sincerely thank the anonymous reviewers for their thorough reviewing and valuable suggestions. The research of Guzman-Olivares was conducted with the support of Bulil Technologies S.L., that provided the hardware for the development of this project, and the financial support of the Spanish Ministry of Science and Innovation, grant PID2022-139131NB-I00. The research of Quijano-Sanchez was conducted with financial support from the Spanish Ministry of Science and Innovation, grants PID2022-139131NB-I00 \& PID2021-122677NB-I00.

\bibliography{sandwich}

\begin{thebibliography}{79}
\providecommand{\natexlab}[1]{#1}

\bibitem[{Balloccu et~al.(2024)Balloccu, Schmidtov{\'a}, Lango, and Dusek}]{leak}
Simone Balloccu, Patr{\'\i}cia Schmidtov{\'a}, Mateusz Lango, and Ondrej Dusek. 2024.
\newblock \href {https://aclanthology.org/2024.eacl-long.5} {Leak, cheat, repeat: Data contamination and evaluation malpractices in closed-source {LLM}s}.
\newblock In \emph{Proceedings of the 18th Conference of the European Chapter of the Association for Computational Linguistics (Volume 1: Long Papers)}, pages 67--93, St. Julian{'}s, Malta. Association for Computational Linguistics.

\bibitem[{Bar-Hillel(1960)}]{BarHillel1960ThePS}
Yehoshua Bar-Hillel. 1960.
\newblock \href {https://aclanthology.org/www.mt-archive.info/50/Bar-Hillel-1960.pdf} {The present status of automatic translation of languages}.
\newblock \emph{Adv. Comput.}, 1:91--163.

\bibitem[{Barba et~al.(2021{\natexlab{a}})Barba, Pasini, and Navigli}]{esc}
Edoardo Barba, Tommaso Pasini, and Roberto Navigli. 2021{\natexlab{a}}.
\newblock \href {https://doi.org/10.18653/v1/2021.naacl-main.371} {{ESC}: Redesigning {WSD} with extractive sense comprehension}.
\newblock In \emph{Proceedings of the 2021 Conference of the North American Chapter of the Association for Computational Linguistics: Human Language Technologies}, pages 4661--4672, Online. Association for Computational Linguistics.

\bibitem[{Barba et~al.(2021{\natexlab{b}})Barba, Procopio, and Navigli}]{consec}
Edoardo Barba, Luigi Procopio, and Roberto Navigli. 2021{\natexlab{b}}.
\newblock \href {https://doi.org/10.18653/v1/2021.emnlp-main.112} {{C}on{S}e{C}: Word sense disambiguation as continuous sense comprehension}.
\newblock In \emph{Proceedings of the 2021 Conference on Empirical Methods in Natural Language Processing}, pages 1492--1503, Online and Punta Cana, Dominican Republic. Association for Computational Linguistics.

\bibitem[{Bevilacqua et~al.(2020)Bevilacqua, Maru, and Navigli}]{bevilacqua-etal-2020-generationary}
Michele Bevilacqua, Marco Maru, and Roberto Navigli. 2020.
\newblock \href {https://doi.org/10.18653/v1/2020.emnlp-main.585} {Generationary or {``}how we went beyond word sense inventories and learned to gloss{''}}.
\newblock In \emph{Proceedings of the 2020 Conference on Empirical Methods in Natural Language Processing (EMNLP)}, pages 7207--7221, Online. Association for Computational Linguistics.

\bibitem[{Bevilacqua and Navigli(2020)}]{ewiser}
Michele Bevilacqua and Roberto Navigli. 2020.
\newblock \href {https://doi.org/10.18653/v1/2020.acl-main.255} {Breaking through the 80{\%} glass ceiling: {R}aising the state of the art in word sense disambiguation by incorporating knowledge graph information}.
\newblock In \emph{Proceedings of the 58th Annual Meeting of the Association for Computational Linguistics}, pages 2854--2864, Online. Association for Computational Linguistics.

\bibitem[{Bevilacqua et~al.(2021)Bevilacqua, Pasini, Raganato, and Navigli}]{wsd_survey}
Michele Bevilacqua, Tommaso Pasini, Alessandro Raganato, and Roberto Navigli. 2021.
\newblock \href {https://doi.org/10.24963/ijcai.2021/593} {Recent trends in word sense disambiguation: A survey}.
\newblock In \emph{Proceedings of the Thirtieth International Joint Conference on Artificial Intelligence, {IJCAI-21}}, pages 4330--4338. International Joint Conferences on Artificial Intelligence Organization.
\newblock Survey Track.

\bibitem[{Blevins and Zettlemoyer(2020)}]{bem}
Terra Blevins and Luke Zettlemoyer. 2020.
\newblock \href {https://doi.org/10.18653/v1/2020.acl-main.95} {Moving down the long tail of word sense disambiguation with gloss informed bi-encoders}.
\newblock In \emph{Proceedings of the 58th Annual Meeting of the Association for Computational Linguistics}, pages 1006--1017, Online. Association for Computational Linguistics.

\bibitem[{Bojanowski et~al.(2017)Bojanowski, Grave, Joulin, and Mikolov}]{fasttext}
Piotr Bojanowski, Edouard Grave, Armand Joulin, and Tomas Mikolov. 2017.
\newblock Enriching word vectors with subword information.
\newblock \emph{Transactions of the Association for Computational Linguistics}, 5:135--146.

\bibitem[{Brown et~al.(2020)Brown, Mann, Ryder, Subbiah, Kaplan, Dhariwal, Neelakantan, Shyam, Sastry, Askell, Agarwal, Herbert-Voss, Krueger, Henighan, Child, Ramesh, Ziegler, Wu, Winter, Hesse, Chen, Sigler, Litwin, Gray, Chess, Clark, Berner, McCandlish, Radford, Sutskever, and Amodei}]{gpt3}
Tom Brown, Benjamin Mann, Nick Ryder, Melanie Subbiah, Jared~D Kaplan, Prafulla Dhariwal, Arvind Neelakantan, Pranav Shyam, Girish Sastry, Amanda Askell, Sandhini Agarwal, Ariel Herbert-Voss, Gretchen Krueger, Tom Henighan, Rewon Child, Aditya Ramesh, Daniel Ziegler, Jeffrey Wu, Clemens Winter, Chris Hesse, Mark Chen, Eric Sigler, Mateusz Litwin, Scott Gray, Benjamin Chess, Jack Clark, Christopher Berner, Sam McCandlish, Alec Radford, Ilya Sutskever, and Dario Amodei. 2020.
\newblock \href {https://proceedings.neurips.cc/paper_files/paper/2020/file/1457c0d6bfcb4967418bfb8ac142f64a-Paper.pdf} {Language models are few-shot learners}.
\newblock In \emph{Advances in Neural Information Processing Systems}, volume~33, pages 1877--1901. Curran Associates, Inc.

\bibitem[{Chiang et~al.(2024)Chiang, Zheng, Sheng, Angelopoulos, Li, Li, Zhang, Zhu, Jordan, Gonzalez, and Stoica}]{chatbot-arena}
Wei-Lin Chiang, Lianmin Zheng, Ying Sheng, Anastasios~Nikolas Angelopoulos, Tianle Li, Dacheng Li, Hao Zhang, Banghua Zhu, Michael Jordan, Joseph~E. Gonzalez, and Ion Stoica. 2024.
\newblock \href {https://arxiv.org/abs/2403.04132} {Chatbot arena: An open platform for evaluating llms by human preference}.
\newblock \emph{Preprint}, arXiv:2403.04132.

\bibitem[{Conia and Navigli(2021)}]{conia-navigli-2021-framing}
Simone Conia and Roberto Navigli. 2021.
\newblock \href {https://doi.org/10.18653/v1/2021.eacl-main.286} {Framing word sense disambiguation as a multi-label problem for model-agnostic knowledge integration}.
\newblock In \emph{Proceedings of the 16th Conference of the European Chapter of the Association for Computational Linguistics: Main Volume}, pages 3269--3275, Online. Association for Computational Linguistics.

\bibitem[{Cowie et~al.(1992)Cowie, Guthrie, and Guthrie}]{cowie-etal-1992-lexical-disambiguation}
Jim Cowie, Joe Guthrie, and Louise Guthrie. 1992.
\newblock \href {https://aclanthology.org/C92-1056} {Lexical disambiguation using simulated annealing}.
\newblock In \emph{{COLING} 1992 Volume 1: The 14th {I}nternational {C}onference on {C}omputational {L}inguistics}.

\bibitem[{Devlin et~al.(2019)Devlin, Chang, Lee, and Toutanova}]{bert}
Jacob Devlin, Ming-Wei Chang, Kenton Lee, and Kristina Toutanova. 2019.
\newblock \href {https://doi.org/10.18653/v1/N19-1423} {{BERT}: Pre-training of deep bidirectional transformers for language understanding}.
\newblock In \emph{Proceedings of the 2019 Conference of the North {A}merican Chapter of the Association for Computational Linguistics: Human Language Technologies, Volume 1 (Long and Short Papers)}, pages 4171--4186, Minneapolis, Minnesota. Association for Computational Linguistics.

\bibitem[{Dong and Sifa(2024)}]{dong-sifa-2024-word}
Tiansi Dong and Rafet Sifa. 2024.
\newblock \href {https://aclanthology.org/2024.neusymbridge-1.3} {Word sense disambiguation as a game of neurosymbolic darts}.
\newblock In \emph{Proceedings of the Workshop: Bridging Neurons and Symbols for Natural Language Processing and Knowledge Graphs Reasoning (NeusymBridge) @ LREC-COLING-2024}, pages 22--32, Torino, Italia. ELRA and ICCL.

\bibitem[{Edmonds and Cotton(2001)}]{edmonds-cotton-2001-senseval}
Philip Edmonds and Scott Cotton. 2001.
\newblock \href {https://aclanthology.org/S01-1001} {{SENSEVAL}-2: Overview}.
\newblock In \emph{Proceedings of {SENSEVAL}-2 Second International Workshop on Evaluating Word Sense Disambiguation Systems}, pages 1--5, Toulouse, France. Association for Computational Linguistics.

\bibitem[{Eisenschlos et~al.(2023)Eisenschlos, Cole, Liu, and Cohen}]{eisenschlos-etal-2023-winodict}
Julian~Martin Eisenschlos, Jeremy~R. Cole, Fangyu Liu, and William~W. Cohen. 2023.
\newblock \href {https://doi.org/10.18653/v1/2023.eacl-main.7} {{W}ino{D}ict: Probing language models for in-context word acquisition}.
\newblock In \emph{Proceedings of the 17th Conference of the European Chapter of the Association for Computational Linguistics}, pages 94--102, Dubrovnik, Croatia. Association for Computational Linguistics.

\bibitem[{Gale et~al.(1992)Gale, Church, and Yarowsky}]{Gale1992AMF}
William~A. Gale, Kenneth~Ward Church, and David Yarowsky. 1992.
\newblock \href {https://api.semanticscholar.org/CorpusID:17567112} {A method for disambiguating word senses in a large corpus}.
\newblock \emph{Computers and the Humanities}, 26:415--439.

\bibitem[{Google(2023)}]{anil2023palm2technicalreport}
Google. 2023.
\newblock \href {https://arxiv.org/abs/2305.10403} {Palm 2 technical report}.
\newblock \emph{Preprint}, arXiv:2305.10403.

\bibitem[{Hadiwinoto et~al.(2019)Hadiwinoto, Ng, and Gan}]{hadiwinoto-etal-2019-improved}
Christian Hadiwinoto, Hwee~Tou Ng, and Wee~Chung Gan. 2019.
\newblock \href {https://doi.org/10.18653/v1/D19-1533} {Improved word sense disambiguation using pre-trained contextualized word representations}.
\newblock In \emph{Proceedings of the 2019 Conference on Empirical Methods in Natural Language Processing and the 9th International Joint Conference on Natural Language Processing (EMNLP-IJCNLP)}, pages 5297--5306, Hong Kong, China. Association for Computational Linguistics.

\bibitem[{Hashimoto and Tsuruoka(2015)}]{hashimoto-tsuruoka-2015-learning}
Kazuma Hashimoto and Yoshimasa Tsuruoka. 2015.
\newblock \href {https://doi.org/10.18653/v1/W15-4001} {Learning embeddings for transitive verb disambiguation by implicit tensor factorization}.
\newblock In \emph{Proceedings of the 3rd Workshop on Continuous Vector Space Models and their Compositionality}, pages 1--11, Beijing, China. Association for Computational Linguistics.

\bibitem[{He et~al.(2020)He, Liu, Gao, and Chen}]{deberta}
Pengcheng He, Xiaodong Liu, Jianfeng Gao, and Weizhu Chen. 2020.
\newblock \href {https://arxiv.org/abs/2006.03654} {Deberta: Decoding-enhanced {BERT} with disentangled attention}.
\newblock \emph{CoRR}, abs/2006.03654.

\bibitem[{Huang et~al.(2019)Huang, Sun, Qiu, and Huang}]{glossbert}
Luyao Huang, Chi Sun, Xipeng Qiu, and Xuanjing Huang. 2019.
\newblock \href {https://doi.org/10.18653/v1/D19-1355} {{G}loss{BERT}: {BERT} for word sense disambiguation with gloss knowledge}.
\newblock In \emph{Proceedings of the 2019 Conference on Empirical Methods in Natural Language Processing and the 9th International Joint Conference on Natural Language Processing (EMNLP-IJCNLP)}, pages 3509--3514, Hong Kong, China. Association for Computational Linguistics.

\bibitem[{Jain and Lobiyal(2015)}]{fuzzy_hindi}
Amita Jain and D.~K. Lobiyal. 2015.
\newblock \href {https://doi.org/10.1145/2790079} {Fuzzy hindi wordnet and word sense disambiguation using fuzzy graph connectivity measures}.
\newblock \emph{ACM Trans. Asian Low-Resour. Lang. Inf. Process.}, 15(2).

\bibitem[{Jiang et~al.(2023)Jiang, Sablayrolles, Mensch, Bamford, Chaplot, de~las Casas, Bressand, Lengyel, Lample, Saulnier, Lavaud, Lachaux, Stock, Scao, Lavril, Wang, Lacroix, and Sayed}]{mistral}
Albert~Q. Jiang, Alexandre Sablayrolles, Arthur Mensch, Chris Bamford, Devendra~Singh Chaplot, Diego de~las Casas, Florian Bressand, Gianna Lengyel, Guillaume Lample, Lucile Saulnier, Lélio~Renard Lavaud, Marie-Anne Lachaux, Pierre Stock, Teven~Le Scao, Thibaut Lavril, Thomas Wang, Timothée Lacroix, and William~El Sayed. 2023.
\newblock \href {https://arxiv.org/abs/2310.06825} {Mistral 7b}.
\newblock \emph{Preprint}, arXiv:2310.06825.

\bibitem[{K{\aa}geb{\"a}ck and Salomonsson(2016)}]{Kgebck2016WordSD}
Mikael K{\aa}geb{\"a}ck and Hans Salomonsson. 2016.
\newblock \href {https://aclanthology.org/W16-5307.pdf} {Word sense disambiguation using a bidirectional lstm}.
\newblock In \emph{CogALex@COLING}.

\bibitem[{Kilgarriff(2004)}]{kilgarriff}
Adam Kilgarriff. 2004.
\newblock \href {https://doi.org/10.1007/978-3-540-30120-2_14} {How dominant is the commonest sense of a word?}
\newblock volume 3206, pages 103--112.

\bibitem[{Kocoń et~al.(2023)Kocoń, Cichecki, Kaszyca, Kochanek, Szydło, Baran, Bielaniewicz, Gruza, Janz, Kanclerz, Kocoń, Koptyra, Mieleszczenko-Kowszewicz, Miłkowski, Oleksy, Piasecki, Łukasz Radliński, Wojtasik, Woźniak, and Kazienko}]{jack-of-all-trades}
Jan Kocoń, Igor Cichecki, Oliwier Kaszyca, Mateusz Kochanek, Dominika Szydło, Joanna Baran, Julita Bielaniewicz, Marcin Gruza, Arkadiusz Janz, Kamil Kanclerz, Anna Kocoń, Bartłomiej Koptyra, Wiktoria Mieleszczenko-Kowszewicz, Piotr Miłkowski, Marcin Oleksy, Maciej Piasecki, Łukasz Radliński, Konrad Wojtasik, Stanisław Woźniak, and Przemysław Kazienko. 2023.
\newblock \href {https://doi.org/10.1016/j.inffus.2023.101861} {Chatgpt: Jack of all trades, master of none}.
\newblock \emph{Information Fusion}, 99:101861.

\bibitem[{Kumar et~al.(2019)Kumar, Jat, Saxena, and Talukdar}]{kumar}
Sawan Kumar, Sharmistha Jat, Karan Saxena, and Partha Talukdar. 2019.
\newblock \href {https://doi.org/10.18653/v1/P19-1568} {Zero-shot word sense disambiguation using sense definition embeddings}.
\newblock In \emph{Proceedings of the 57th Annual Meeting of the Association for Computational Linguistics}, pages 5670--5681, Florence, Italy. Association for Computational Linguistics.

\bibitem[{Lesk(1986)}]{lesk}
Michael Lesk. 1986.
\newblock \href {https://doi.org/10.1145/318723.318728} {Automatic sense disambiguation using machine readable dictionaries: how to tell a pine cone from an ice cream cone}.
\newblock In \emph{Proceedings of the 5th Annual International Conference on Systems Documentation}, SIGDOC '86, page 24–26, New York, NY, USA. Association for Computing Machinery.

\bibitem[{Lewis et~al.(2020)Lewis, Liu, Goyal, Ghazvininejad, Mohamed, Levy, Stoyanov, and Zettlemoyer}]{lewis-etal-2020-bart}
Mike Lewis, Yinhan Liu, Naman Goyal, Marjan Ghazvininejad, Abdelrahman Mohamed, Omer Levy, Veselin Stoyanov, and Luke Zettlemoyer. 2020.
\newblock \href {https://doi.org/10.18653/v1/2020.acl-main.703} {{BART}: Denoising sequence-to-sequence pre-training for natural language generation, translation, and comprehension}.
\newblock In \emph{Proceedings of the 58th Annual Meeting of the Association for Computational Linguistics}, pages 7871--7880, Online. Association for Computational Linguistics.

\bibitem[{Liu et~al.(2023)Liu, Ning, Teng, Liu, Zhou, and Zhang}]{liu2023evaluating}
Hanmeng Liu, Ruoxi Ning, Zhiyang Teng, Jian Liu, Qiji Zhou, and Yue Zhang. 2023.
\newblock \href {https://arxiv.org/abs/2304.03439} {Evaluating the logical reasoning ability of chatgpt and gpt-4}.
\newblock \emph{Preprint}, arXiv:2304.03439.

\bibitem[{Liu and Zeng(2024)}]{dhfm}
Jiaheng Liu and Haonan Zeng. 2024.
\newblock \href {https://doi.org/10.1145/3652628.3652691} {Achieving fine-grained word sense disambiguation with context hypergraph and sememe hypergraph}.
\newblock In \emph{Proceedings of the 4th International Conference on Artificial Intelligence and Computer Engineering}, ICAICE '23, page 383–388, New York, NY, USA. Association for Computing Machinery.

\bibitem[{Liu et~al.(2020)Liu, Gu, Goyal, Li, Edunov, Ghazvininejad, Lewis, and Zettlemoyer}]{mbart}
Yinhan Liu, Jiatao Gu, Naman Goyal, Xian Li, Sergey Edunov, Marjan Ghazvininejad, Mike Lewis, and Luke Zettlemoyer. 2020.
\newblock \href {https://doi.org/10.1162/tacl_a_00343} {Multilingual denoising pre-training for neural machine translation}.
\newblock \emph{Transactions of the Association for Computational Linguistics}, 8:726--742.

\bibitem[{Liu et~al.(2019)Liu, Ott, Goyal, Du, Joshi, Chen, Levy, Lewis, Zettlemoyer, and Stoyanov}]{liu2019robertarobustlyoptimizedbert}
Yinhan Liu, Myle Ott, Naman Goyal, Jingfei Du, Mandar Joshi, Danqi Chen, Omer Levy, Mike Lewis, Luke Zettlemoyer, and Veselin Stoyanov. 2019.
\newblock \href {https://arxiv.org/abs/1907.11692} {Roberta: A robustly optimized bert pretraining approach}.
\newblock \emph{Preprint}, arXiv:1907.11692.

\bibitem[{Liu and Liu(2023)}]{liu-liu-2023-ambiguity}
Zhu Liu and Ying Liu. 2023.
\newblock \href {https://doi.org/10.18653/v1/2023.findings-acl.245} {Ambiguity meets uncertainty: Investigating uncertainty estimation for word sense disambiguation}.
\newblock In \emph{Findings of the Association for Computational Linguistics: ACL 2023}, pages 3963--3977, Toronto, Canada. Association for Computational Linguistics.

\bibitem[{Loshchilov and Hutter(2017)}]{loshchilov2017sgdrstochasticgradientdescent}
Ilya Loshchilov and Frank Hutter. 2017.
\newblock \href {https://arxiv.org/abs/1608.03983} {Sgdr: Stochastic gradient descent with warm restarts}.
\newblock \emph{Preprint}, arXiv:1608.03983.

\bibitem[{Loureiro and Jorge(2019)}]{loureiro-jorge-2019-language}
Daniel Loureiro and Al{\'\i}pio Jorge. 2019.
\newblock \href {https://doi.org/10.18653/v1/P19-1569} {Language modelling makes sense: Propagating representations through {W}ord{N}et for full-coverage word sense disambiguation}.
\newblock In \emph{Proceedings of the 57th Annual Meeting of the Association for Computational Linguistics}, pages 5682--5691, Florence, Italy. Association for Computational Linguistics.

\bibitem[{Luo et~al.(2018)Luo, Liu, Xia, Chang, and Sui}]{luo-etal-2018-incorporating}
Fuli Luo, Tianyu Liu, Qiaolin Xia, Baobao Chang, and Zhifang Sui. 2018.
\newblock \href {https://doi.org/10.18653/v1/P18-1230} {Incorporating glosses into neural word sense disambiguation}.
\newblock In \emph{Proceedings of the 56th Annual Meeting of the Association for Computational Linguistics (Volume 1: Long Papers)}, pages 2473--2482, Melbourne, Australia. Association for Computational Linguistics.

\bibitem[{Mac~Lane and Birkhoff(1967)}]{key0214415m}
Saunders Mac~Lane and Garrett Birkhoff. 1967.
\newblock \emph{Algebra}.
\newblock Macmillan, New York.
\newblock MR:0214415. Zbl:0153.32401.

\bibitem[{Maru et~al.(2022)Maru, Conia, Bevilacqua, and Navigli}]{maru-etal-2022-nibbling}
Marco Maru, Simone Conia, Michele Bevilacqua, and Roberto Navigli. 2022.
\newblock \href {https://doi.org/10.18653/v1/2022.acl-long.324} {{N}ibbling at the hard core of {W}ord {S}ense {D}isambiguation}.
\newblock In \emph{Proceedings of the 60th Annual Meeting of the Association for Computational Linguistics (Volume 1: Long Papers)}, pages 4724--4737, Dublin, Ireland. Association for Computational Linguistics.

\bibitem[{McCarthy et~al.(2016)McCarthy, Apidianaki, and Erk}]{mccarthy-etal-2016-word}
Diana McCarthy, Marianna Apidianaki, and Katrin Erk. 2016.
\newblock \href {https://doi.org/10.1162/COLI_a_00247} {Word sense clustering and clusterability}.
\newblock \emph{Computational Linguistics}, 42(2):245--275.

\bibitem[{Mikolov et~al.(2013)Mikolov, Chen, Corrado, and Dean}]{word2vec}
Tomas Mikolov, Kai Chen, G.s Corrado, and Jeffrey Dean. 2013.
\newblock Efficient estimation of word representations in vector space.
\newblock \emph{Proceedings of Workshop at ICLR}, 2013.

\bibitem[{Miller et~al.(1993)Miller, Leacock, Tengi, and Bunker}]{miller}
George Miller, Claudia Leacock, Randee Tengi, and Ross Bunker. 1993.
\newblock \href {https://doi.org/10.3115/1075671.1075742} {A semantic concordance}.
\newblock pages 303--308.

\bibitem[{Minaee et~al.(2024)Minaee, Mikolov, Nikzad, Chenaghlu, Socher, Amatriain, and Gao}]{minaee-sruvey}
Shervin Minaee, Tomas Mikolov, Narjes Nikzad, Meysam Chenaghlu, Richard Socher, Xavier Amatriain, and Jianfeng Gao. 2024.
\newblock \href {https://arxiv.org/abs/2402.06196} {Large language models: A survey}.
\newblock \emph{Preprint}, arXiv:2402.06196.

\bibitem[{Mizuki and Okazaki(2023)}]{mizuki-okazaki-2023-semantic}
Sakae Mizuki and Naoaki Okazaki. 2023.
\newblock \href {https://doi.org/10.18653/v1/2023.eacl-main.251} {Semantic specialization for knowledge-based word sense disambiguation}.
\newblock In \emph{Proceedings of the 17th Conference of the European Chapter of the Association for Computational Linguistics}, pages 3457--3470, Dubrovnik, Croatia. Association for Computational Linguistics.

\bibitem[{Moro and Navigli(2015)}]{moro-navigli-2015-semeval}
Andrea Moro and Roberto Navigli. 2015.
\newblock \href {https://doi.org/10.18653/v1/S15-2049} {{S}em{E}val-2015 task 13: Multilingual all-words sense disambiguation and entity linking}.
\newblock In \emph{Proceedings of the 9th International Workshop on Semantic Evaluation ({S}em{E}val 2015)}, pages 288--297, Denver, Colorado. Association for Computational Linguistics.

\bibitem[{Moro et~al.(2014)Moro, Raganato, and Navigli}]{moro-etal-2014-entity}
Andrea Moro, Alessandro Raganato, and Roberto Navigli. 2014.
\newblock \href {https://doi.org/10.1162/tacl_a_00179} {Entity linking meets word sense disambiguation: a unified approach}.
\newblock \emph{Transactions of the Association for Computational Linguistics}, 2:231--244.

\bibitem[{Navigli et~al.(2023)Navigli, Conia, and Ross}]{bias}
Roberto Navigli, Simone Conia, and Bj\"{o}rn Ross. 2023.
\newblock \href {https://doi.org/10.1145/3597307} {Biases in large language models: Origins, inventory, and discussion}.
\newblock \emph{J. Data and Information Quality}, 15(2).

\bibitem[{Navigli et~al.(2013)Navigli, Jurgens, and Vannella}]{navigli-etal-2013-semeval}
Roberto Navigli, David Jurgens, and Daniele Vannella. 2013.
\newblock \href {https://aclanthology.org/S13-2040} {{S}em{E}val-2013 task 12: Multilingual word sense disambiguation}.
\newblock In \emph{Second Joint Conference on Lexical and Computational Semantics (*{SEM}), Volume 2: Proceedings of the Seventh International Workshop on Semantic Evaluation ({S}em{E}val 2013)}, pages 222--231, Atlanta, Georgia, USA. Association for Computational Linguistics.

\bibitem[{Navigli and Ponzetto(2010)}]{navigli-ponzetto-2010-babelnet}
Roberto Navigli and Simone~Paolo Ponzetto. 2010.
\newblock \href {https://aclanthology.org/P10-1023} {{B}abel{N}et: Building a very large multilingual semantic network}.
\newblock In \emph{Proceedings of the 48th Annual Meeting of the Association for Computational Linguistics}, pages 216--225, Uppsala, Sweden. Association for Computational Linguistics.

\bibitem[{OpenAI(2023)}]{gpt4}
OpenAI. 2023.
\newblock \href {https://api.semanticscholar.org/CorpusID:257532815} {Gpt-4 technical report}.

\bibitem[{Ouyang et~al.(2022)Ouyang, Wu, Jiang, Almeida, Wainwright, Mishkin, Zhang, Agarwal, Slama, Gray, Schulman, Hilton, Kelton, Miller, Simens, Askell, Welinder, Christiano, Leike, and Lowe}]{instructgpt}
Long Ouyang, Jeffrey Wu, Xu~Jiang, Diogo Almeida, Carroll Wainwright, Pamela Mishkin, Chong Zhang, Sandhini Agarwal, Katarina Slama, Alex Gray, John Schulman, Jacob Hilton, Fraser Kelton, Luke Miller, Maddie Simens, Amanda Askell, Peter Welinder, Paul Christiano, Jan Leike, and Ryan Lowe. 2022.
\newblock \href {https://proceedings.neurips.cc/paper_files/paper/2022/file/b1efde53be364a73914f58805a001731-Paper-Conference.pdf} {Training language models to follow instructions with human feedback}.
\newblock In \emph{Advances in Neural Information Processing Systems}.

\bibitem[{Pasini et~al.(2021)Pasini, Raganato, and Navigli}]{Pasini_Raganato_Navigli_2021}
Tommaso Pasini, Alessandro Raganato, and Roberto Navigli. 2021.
\newblock \href {https://doi.org/10.1609/aaai.v35i15.17609} {Xl-wsd: An extra-large and cross-lingual evaluation framework for word sense disambiguation}.
\newblock \emph{Proceedings of the AAAI Conference on Artificial Intelligence}, 35(15):13648--13656.

\bibitem[{Penedo et~al.(2023)Penedo, Malartic, Hesslow, Cojocaru, Alobeidli, Cappelli, Pannier, Almazrouei, and Launay}]{falcon}
Guilherme Penedo, Quentin Malartic, Daniel Hesslow, Ruxandra Cojocaru, Hamza Alobeidli, Alessandro Cappelli, Baptiste Pannier, Ebtesam Almazrouei, and Julien Launay. 2023.
\newblock \href {https://proceedings.neurips.cc/paper_files/paper/2023/file/fa3ed726cc5073b9c31e3e49a807789c-Paper-Datasets_and_Benchmarks.pdf} {The refinedweb dataset for falcon llm: Outperforming curated corpora with web data only}.
\newblock In \emph{Advances in Neural Information Processing Systems}, volume~36, pages 79155--79172. Curran Associates, Inc.

\bibitem[{Pennington et~al.(2014)Pennington, Socher, and Manning}]{glove}
Jeffrey Pennington, Richard Socher, and Christopher Manning. 2014.
\newblock \href {https://doi.org/10.3115/v1/D14-1162} {{G}lo{V}e: Global vectors for word representation}.
\newblock In \emph{Proceedings of the 2014 Conference on Empirical Methods in Natural Language Processing ({EMNLP})}, pages 1532--1543, Doha, Qatar. Association for Computational Linguistics.

\bibitem[{Pradhan et~al.(2007)Pradhan, Loper, Dligach, and Palmer}]{pradhan-etal-2007-semeval}
Sameer Pradhan, Edward Loper, Dmitriy Dligach, and Martha Palmer. 2007.
\newblock \href {https://aclanthology.org/S07-1016} {{S}em{E}val-2007 task-17: {E}nglish lexical sample, {SRL} and all words}.
\newblock In \emph{Proceedings of the Fourth International Workshop on Semantic Evaluations ({S}em{E}val-2007)}, pages 87--92, Prague, Czech Republic. Association for Computational Linguistics.

\bibitem[{Qin et~al.(2023)Qin, Zhang, Zhang, Chen, Yasunaga, and Yang}]{chatgpt-reason}
Chengwei Qin, Aston Zhang, Zhuosheng Zhang, Jiaao Chen, Michihiro Yasunaga, and Diyi Yang. 2023.
\newblock \href {https://doi.org/10.18653/v1/2023.emnlp-main.85} {Is {C}hat{GPT} a general-purpose natural language processing task solver?}
\newblock In \emph{Proceedings of the 2023 Conference on Empirical Methods in Natural Language Processing}, pages 1339--1384, Singapore. Association for Computational Linguistics.

\bibitem[{Qorib et~al.(2024)Qorib, Moon, and Ng}]{qorib-etal-2024-decoder}
Muhammad Qorib, Geonsik Moon, and Hwee~Tou Ng. 2024.
\newblock \href {https://doi.org/10.18653/v1/2024.findings-acl.967} {Are decoder-only language models better than encoder-only language models in understanding word meaning?}
\newblock In \emph{Findings of the Association for Computational Linguistics ACL 2024}, pages 16339--16347, Bangkok, Thailand and virtual meeting. Association for Computational Linguistics.

\bibitem[{Raganato et~al.(2017{\natexlab{a}})Raganato, Camacho-Collados, and Navigli}]{raganato-etal-2017-word}
Alessandro Raganato, Jose Camacho-Collados, and Roberto Navigli. 2017{\natexlab{a}}.
\newblock \href {https://aclanthology.org/E17-1010} {Word sense disambiguation: A unified evaluation framework and empirical comparison}.
\newblock In \emph{Proceedings of the 15th Conference of the {E}uropean Chapter of the Association for Computational Linguistics: Volume 1, Long Papers}, pages 99--110, Valencia, Spain. Association for Computational Linguistics.

\bibitem[{Raganato et~al.(2017{\natexlab{b}})Raganato, Delli~Bovi, and Navigli}]{raganato-etal-2017-neural}
Alessandro Raganato, Claudio Delli~Bovi, and Roberto Navigli. 2017{\natexlab{b}}.
\newblock \href {https://doi.org/10.18653/v1/D17-1120} {Neural sequence learning models for word sense disambiguation}.
\newblock In \emph{Proceedings of the 2017 Conference on Empirical Methods in Natural Language Processing}, pages 1156--1167, Copenhagen, Denmark. Association for Computational Linguistics.

\bibitem[{Reimers and Gurevych(2019)}]{reimers-gurevych-2019-sentence}
Nils Reimers and Iryna Gurevych. 2019.
\newblock \href {https://doi.org/10.18653/v1/D19-1410} {Sentence-{BERT}: Sentence embeddings using {S}iamese {BERT}-networks}.
\newblock In \emph{Proceedings of the 2019 Conference on Empirical Methods in Natural Language Processing and the 9th International Joint Conference on Natural Language Processing (EMNLP-IJCNLP)}, pages 3982--3992, Hong Kong, China. Association for Computational Linguistics.

\bibitem[{Rosso et~al.(2005)Rosso, Montes-y G\'{o}mez, Buscaldi, Pancardo-Rodr\'{\i}guez, and Pineda}]{nouns}
Paolo Rosso, Manuel Montes-y G\'{o}mez, Davide Buscaldi, Aar\'{o}n Pancardo-Rodr\'{\i}guez, and Luis Villase\~{n}or Pineda. 2005.
\newblock \href {https://doi.org/10.1007/978-3-540-30586-6_30} {Two web-based approaches for noun sense disambiguation}.
\newblock In \emph{Proceedings of the 6th International Conference on Computational Linguistics and Intelligent Text Processing}, CICLing'05, page 267–279, Berlin, Heidelberg. Springer-Verlag.

\bibitem[{Scarlini et~al.(2020)Scarlini, Pasini, and Navigli}]{scarlini-etal-2020-contexts}
Bianca Scarlini, Tommaso Pasini, and Roberto Navigli. 2020.
\newblock \href {https://doi.org/10.18653/v1/2020.emnlp-main.285} {With more contexts comes better performance: Contextualized sense embeddings for all-round word sense disambiguation}.
\newblock In \emph{Proceedings of the 2020 Conference on Empirical Methods in Natural Language Processing (EMNLP)}, pages 3528--3539, Online. Association for Computational Linguistics.

\bibitem[{Scozzafava et~al.(2020)Scozzafava, Maru, Brignone, Torrisi, and Navigli}]{scozzafava-etal-2020-personalized}
Federico Scozzafava, Marco Maru, Fabrizio Brignone, Giovanni Torrisi, and Roberto Navigli. 2020.
\newblock \href {https://doi.org/10.18653/v1/2020.acl-demos.6} {Personalized {P}age{R}ank with syntagmatic information for multilingual word sense disambiguation}.
\newblock In \emph{Proceedings of the 58th Annual Meeting of the Association for Computational Linguistics: System Demonstrations}, pages 37--46, Online. Association for Computational Linguistics.

\bibitem[{Snyder and Palmer(2004)}]{snyder-palmer-2004-english}
Benjamin Snyder and Martha Palmer. 2004.
\newblock \href {https://aclanthology.org/W04-0811} {The {E}nglish all-words task}.
\newblock In \emph{Proceedings of {SENSEVAL}-3, the Third International Workshop on the Evaluation of Systems for the Semantic Analysis of Text}, pages 41--43, Barcelona, Spain. Association for Computational Linguistics.

\bibitem[{Song et~al.(2021)Song, Ong, Ng, and Lin}]{song-etal-2021-improved-word}
Yang Song, Xin~Cai Ong, Hwee~Tou Ng, and Qian Lin. 2021.
\newblock \href {https://doi.org/10.18653/v1/2021.findings-emnlp.365} {Improved word sense disambiguation with enhanced sense representations}.
\newblock In \emph{Findings of the Association for Computational Linguistics: EMNLP 2021}, pages 4311--4320, Punta Cana, Dominican Republic. Association for Computational Linguistics.

\bibitem[{Sumanathilaka et~al.(2024)Sumanathilaka, Micallef, and Hough}]{10691283}
Deshan Sumanathilaka, Nicholas Micallef, and Julian Hough. 2024.
\newblock \href {https://doi.org/10.1109/ICSGRC62081.2024.10691283} {Assessing gpt's potential for word sense disambiguation: A quantitative evaluation on prompt engineering techniques}.
\newblock In \emph{2024 IEEE 15th Control and System Graduate Research Colloquium (ICSGRC)}, pages 204--209.

\bibitem[{Taghipour and Ng(2015)}]{taghipour-ng-2015-semi}
Kaveh Taghipour and Hwee~Tou Ng. 2015.
\newblock \href {https://doi.org/10.3115/v1/N15-1035} {Semi-supervised word sense disambiguation using word embeddings in general and specific domains}.
\newblock In \emph{Proceedings of the 2015 Conference of the North {A}merican Chapter of the Association for Computational Linguistics: Human Language Technologies}, pages 314--323, Denver, Colorado. Association for Computational Linguistics.

\bibitem[{Touvron et~al.(2023)Touvron, Lavril, Izacard, Martinet, Lachaux, Lacroix, Rozière, Goyal, Hambro, Azhar, Rodriguez, Joulin, Grave, and Lample}]{llama}
Hugo Touvron, Thibaut Lavril, Gautier Izacard, Xavier Martinet, Marie-Anne Lachaux, Timothée Lacroix, Baptiste Rozière, Naman Goyal, Eric Hambro, Faisal Azhar, Aurelien Rodriguez, Armand Joulin, Edouard Grave, and Guillaume Lample. 2023.
\newblock \href {https://arxiv.org/abs/2302.13971} {Llama: Open and efficient foundation language models}.
\newblock \emph{Preprint}, arXiv:2302.13971.

\bibitem[{Vaswani et~al.(2017)Vaswani, Shazeer, Parmar, Uszkoreit, Jones, Gomez, Kaiser, and Polosukhin}]{attention}
Ashish Vaswani, Noam Shazeer, Niki Parmar, Jakob Uszkoreit, Llion Jones, Aidan~N Gomez, \L~ukasz Kaiser, and Illia Polosukhin. 2017.
\newblock \href {https://proceedings.neurips.cc/paper_files/paper/2017/file/3f5ee243547dee91fbd053c1c4a845aa-Paper.pdf} {Attention is all you need}.
\newblock In \emph{Advances in Neural Information Processing Systems}, volume~30. Curran Associates, Inc.

\bibitem[{Vial et~al.(2019)Vial, Lecouteux, and Schwab}]{vial-etal-2019-sense}
Lo{\"\i}c Vial, Benjamin Lecouteux, and Didier Schwab. 2019.
\newblock \href {https://aclanthology.org/2019.gwc-1.14} {Sense vocabulary compression through the semantic knowledge of {W}ord{N}et for neural word sense disambiguation}.
\newblock In \emph{Proceedings of the 10th Global Wordnet Conference}, pages 108--117, Wroclaw, Poland. Global Wordnet Association.

\bibitem[{Wagner(2009)}]{Wagner2009VerbSD}
Wiebke Wagner. 2009.
\newblock \href {https://api.semanticscholar.org/CorpusID:3110686} {Verb sense disambiguation using a predicate-argument-clustering model}.

\bibitem[{Wang et~al.(2015)Wang, Bansal, Gimpel, Ziebart, and Yu}]{wang-etal-2015-sense}
Jing Wang, Mohit Bansal, Kevin Gimpel, Brian~D. Ziebart, and Clement~T. Yu. 2015.
\newblock \href {https://doi.org/10.1162/tacl_a_00122} {A sense-topic model for word sense induction with unsupervised data enrichment}.
\newblock \emph{Transactions of the Association for Computational Linguistics}, 3:59--71.

\bibitem[{Weaver(1949/1955)}]{weaver}
Warren Weaver. 1949/1955.
\newblock Translation.
\newblock In William~N. Locke and A.~Donald Boothe, editors, \emph{Machine Translation of Languages}, pages 15--23. MIT Press, Cambridge, MA.
\newblock Reprinted from a memorandum written by Weaver in 1949.

\bibitem[{Yarowsky(1992)}]{Yarowsky1992WordSenseDU}
David Yarowsky. 1992.
\newblock \href {https://api.semanticscholar.org/CorpusID:1693468} {Word-sense disambiguation using statistical models of roget's categories trained on large corpora}.
\newblock In \emph{International Conference on Computational Linguistics}.

\bibitem[{Yarowsky(1995)}]{Yarowsky1995UnsupervisedWS}
David Yarowsky. 1995.
\newblock \href {https://aclanthology.org/P95-1026.pdf} {Unsupervised word sense disambiguation rivaling supervised methods}.
\newblock In \emph{Annual Meeting of the Association for Computational Linguistics}.

\bibitem[{Yuan et~al.(2016)Yuan, Richardson, Doherty, Evans, and Altendorf}]{yuan-etal-2016-semi}
Dayu Yuan, Julian Richardson, Ryan Doherty, Colin Evans, and Eric Altendorf. 2016.
\newblock \href {https://aclanthology.org/C16-1130} {Semi-supervised word sense disambiguation with neural models}.
\newblock In \emph{Proceedings of {COLING} 2016, the 26th International Conference on Computational Linguistics: Technical Papers}, pages 1374--1385, Osaka, Japan. The COLING 2016 Organizing Committee.

\bibitem[{Zhang et~al.(2023)Zhang, He, and Guo}]{zhang}
Junwei Zhang, Ruifang He, and Fengyu Guo. 2023.
\newblock \href {https://doi.org/10.1609/aaai.v37i11.26633} {Quantum-inspired representation for long-tail senses of word sense disambiguation}.
\newblock In \emph{Proceedings of the Thirty-Seventh AAAI Conference on Artificial Intelligence and Thirty-Fifth Conference on Innovative Applications of Artificial Intelligence and Thirteenth Symposium on Educational Advances in Artificial Intelligence}, AAAI'23/IAAI'23/EAAI'23. AAAI Press.

\end{thebibliography}

\appendix
\section{Additional Implementation details}
\label{apdx:training}
\subsection{Hardware Specifications}
All experiments were performed in a machine with the technical capabilities reported in \cref{tab:resources}.
\begin{table}[h!]
\begin{tabular}{|l|l|}
\hline
CPU   & AMD Ryzen Threadripper 3975WX \\ \hline
RAM   & 256 GB                            \\ \hline
Cores & 64                                \\ \hline
GPU   & 2x Nvidia A100 160GB                   \\ \hline
\end{tabular}
\label{tab:resources}
\caption{Specifications of the machine in which the experiments were executed.}
\end{table}

\subsection{Training Hyperparameters}

The full table of hyperparameters used in the training of the system can be found in \cref{table: hyperparams}. Different options for the settings of the system appear between curly braces, while the selected ones appear in bold. The only hyperparameter endemic to the SANDWiCH system is the number K of candidate senses returned in the coarse search (see \cref{fig:architecture}).
\begin{table*}
\centering
\begin{tabular}{ll}
\hline
\textbf{Parameter}                   & \textbf{Value}                      \\ \hline
Optimizer                   & AdamW                      \\
Learning Rate               & \{$10^{-7}$, $10^{-6}, \bm{10^{-5}}, 10^{-4}$\}\\
Gradient Accumulation Steps & \{1,\textbf{5},10\}                          \\
Maximum Gradient Norm       & \{\textbf{1}, 5, 10, 50, 100\}                          \\
Batch Size                  & \{4, 16, 32, \textbf{64}, 128\}                         \\
Epochs                      & {1, 5, \textbf{10}, 15, 20}                         \\
Evaluation Steps            & 1000                       \\
Scheduler                   & \{\textbf{Cosine Annealing}, Linear\}           \\
Weight Decay                & 0.01                       \\
Maximum Gradient Norm       & {\textbf{1}, 5, 10}                          \\
Loss Function               & Cross-Entropy with logits  \\
Max Tokens                  & 512\\
K (Top K retrieval SANDWiCH)                          & {5, 10, 15, 20, 25, \textbf{30}, 35, 40}\\   
\end{tabular}
\caption{Training hyperparameters for the proposed system. Between curly braces are all values tested during optimization, the one selected are marked in bold.}
\label{table: hyperparams}
\end{table*}

\subsection{Transforming from WordNet Synset to BabelNet Synsets}
Our system uses a dump of BabelNet 5.0 as its information source. The graph we employ is a post-processed version that is restricted to a specific language. Given that we work with the version from the \citet{raganato-etal-2017-neural} dataset, implemented by \citet{Pasini_Raganato_Navigli_2021}, for the all-English WSD task, we had to adapt our comparison across models to accommodate BabelNet. This involves mapping the WordNet-based results of some systems (like ConSec) to BabelNet synsets. Since WordNet differentiates synsets at a finer level, we adjust the predictions from WordNet-based systems by associating all related BabelNet synsets to the predicted WordNet synset. These are then treated as a single unit when compared to the gold standard group. If there is any overlap between the predicted and the gold standard synsets, the prediction is considered correct. To ensure the accuracy of our comparison method, we reproduced all results reported in the ConSec paper, validating the correctness of our mapping methodology and ensuring that no system has an unfair advantage.
\begin{figure*}[ht!]
    \centering
    \includegraphics[width=0.7\linewidth]{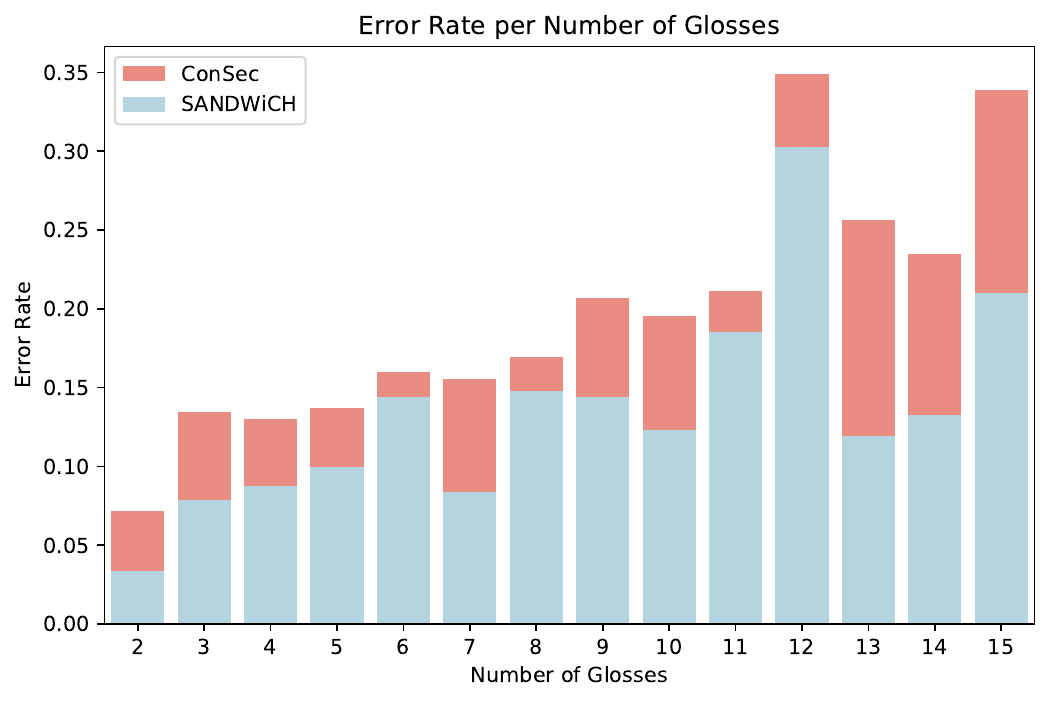}
    \caption{Error rate difference between ConSec (in salmon) and SANDWiCH (in light blue) for words with different number of glosses.}
    \label{fig:error-rate-poli}
\end{figure*}
\section{Polysemic Words accuracy comparison}
\label{apdx:polysemic}
In this section, we compare the performance of ConSec, the previous state-of-the-art model, with the SANDWiCH framework on the all-words English WSD dataset, focusing on polysemic words grouped by their number of possible meanings (see \cref{fig:error-rate-poli}). SANDWiCH consistently reduces the error across all polysemic words, with this reduction becoming more pronounced as the number of possible senses increases. This suggests that the clustering approach employed by SANDWiCH is more effective in managing words with multiple senses and is less dependent on the frequency with which a particular sense appears in the training data.

\section{Licensing and BabelNet derived data}
BabelNet is covered under a license that does not permit the usage of the resource or any derived products from it for other purposes than scientific research. For this reason, following the terms stated in BabelNet's license, we explicity prohibit the usage of the derived sense networks or the model trained with them for any usage different than scientific research.

\end{document}